\documentclass[sn-mathphys,Numbered]{sn-jnl}% Math and Physical Sciences Reference Style
\usepackage{graphicx}%
\usepackage{multirow}%
\usepackage{amsmath,amssymb,amsfonts}%
\usepackage{amsthm}%
\usepackage{mathrsfs}%
\usepackage[title]{appendix}%
\usepackage{xcolor}%
\usepackage{textcomp}%
\usepackage{manyfoot}%
\usepackage{booktabs}%
\usepackage{algorithm}%
\usepackage{algorithmicx}%
\usepackage{algpseudocode}%
\usepackage{listings}%
\theoremstyle{thmstyleone}%
\theoremstyle{thmstyletwo}%

\theoremstyle{thmstylethree}%

\begin{document}

\title[Virtual heatmaps to optimize point of entry location in lung biopsy planning systems]{Virtual airways heatmaps to optimize point of entry location in lung biopsy planning systems}

%%=============================================================%%
%% Prefix	-> \pfx{Dr}
%% GivenName	-> \fnm{Joergen W.}
%% Particle	-> \spfx{van der} -> surname prefix
%% FamilyName	-> \sur{Ploeg}
%% Suffix	-> \sfx{IV}
%% NatureName	-> \tanm{Poet Laureate} -> Title after name
%% Degrees	-> \dgr{MSc, PhD}
%% \author*[1,2]{\pfx{Dr} \fnm{Joergen W.} \spfx{van der} \sur{Ploeg} \sfx{IV} \tanm{Poet Laureate} 
%%                 \dgr{MSc, PhD}}\email{iauthor@gmail.com}
%%=============================================================%%

\author*[1]{\fnm{Debora} \sur{Gil}}\email{debora@cvc.uab.cat}
\author[1]{\fnm{Pere} \sur{Lloret}}\email{plloret@cvc.uab.cat}
\author[2]{\fnm{Marta} \sur{Diez-Ferrer}}\email{marta.diez@bellvitgehospital.cat}
\author*[1]{\fnm{Carles} \sur{Sanchez}}\email{csanchez@cvc.uab.cat}

\affil*[1]{\orgdiv{Computer Science Department}, \orgname{Universitat Aut\`onoma Barcelona and Computer Vision Center}}

\affil[2]{\orgdiv{Department of Respiratory Medicine,}, \orgname{Hospital Universitari de Bellvitge}, \orgaddress{\city{L’Hospitalet de Llobregat}}}

%%==================================%%
%% sample for unstructured abstract %%
%%==================================%%

\abstract{\textbf{Purpose:} We present a virtual model to optimize point of entry (POE) in lung biopsy planning systems. Our model allows to compute the quality of a biopsy sample taken from potential POE, taking into account the margin of error that arises from discrepancies between the orientation in the planning simulation and the actual orientation during the operation. Additionally, the study examines the impact of the characteristics of the lesión. 

\textbf{Methods:} The quality of the biopsy is given by a heatmap projected onto the skeleton of a patient-specific model of airways. The skeleton provides a 3D representation of airways structure, while the heatmap intensity represents the potential amount of tissue that it could be extracted from each POE. This amount of tissue is determined by the intersection of the lesion with a cone that represents the uncertainty area in the introduction of biopsy instruments.  The cone, lesion, and skeleton are modelled as graphical objects that define a 3D scene of the intervention.

\textbf{Results:} We have simulated different settings of the intervention scene from a single anatomy extracted from a CT scan and two lesions with regular and irregular shapes. The different scenarios are simulated by systematic rotation of each lesion placed at different distances from airways. Analysis of the heatmaps for the different settings show a strong impact of lesion orientation for irregular shape and the distance for both shapes.

\textbf{Conclusion:} The proposed heatmaps help to visually assess the optimal POE and identify whether multiple optimal POEs exist in different zones of the bronchi. They also allow us to model the maximum allowable error in navigation systems and study which variables have the greatest influence on the success of the operation. Additionally, they help determine at what point this influence could potentially jeopardize the operation.
}

\keywords{Biopsy Feasibility, Virtual Bronchoscopy, Point of Entry, Heatmaps}

%%\pacs[JEL Classification]{D8, H51}

%%\pacs[MSC Classification]{35A01, 65L10, 65L12, 65L20, 65L70}

\maketitle

\section{Introduction}\label{sec1}

%Cancer is the leading cause of death worldwide qith approximately 1 in 6 deaths due to this disease. Lung cancer specifically, is the second most common type of cancer that exists \cite{jemal2011global}. Cancer is detected by performing a biopsy, an operation through which a sample of the tissue to be analyzed is extracted. One of the most used methodologies to extract tissue samples in lung cancer is to perform a bronchoscopy. A bronchoscopy is a medical procedure in which a thin, flexible tube with a camera at the end is inserted through the mouth or nose to examine the airways and lungs. 

Bronchoscopy is a standard procedure to extract lung tissue samples for histopathological analysis. Although there are currently various technologies—such as endobronchial ultrasound and navigation—that enable guiding the doctor to the lesion, the reality is that in most centers where diagnostic bronchoscopy is performed, these technologies are not available as its expensiveness and expertness \cite{nadig2023guided}. Conventional bronchoscopy may be the only bronchoscopic option available in centers without interventional expertise \cite{shepherd2018image}.

Virtual bronchoscopy (VB) is the usual approach to computational planning. Virtual bronchoscopies combine medical imaging, 3D visualization technologies and simulations to enable a virtual exploration of 3D reconstruction of airways extracted from CT scans. 
These systems provide doctors with several POE  with optimal distance to the lesion and instrumental orientation. During the procedure, the pulmonologist must replicate the previously planned navigation. The guidance systems \cite{ramzy2018biopsy,burks2022electromagnetic,avasarala2022sight} help the doctor to navigate through the bronchi following the planned route, positioning and orienting the bronchoscope, according to the virtual biopsy. 

Regardless of the guidance technology, guidance systems have a margin of error in, both, location of the POE and orientation of sampling and tunneling instruments. These errors might have a different impact on the quality of the sample depending on specific anatomical and morphological conditions, like lesion shape irregularity, distance and orientation to the POE and lesion size, among others \cite{diez2019ultrathin}.  The existing planning systems provide position and orientation, but do not incorporate the feasibility of the intervention taking into account the errors of the navigation system or the patient anatomy \cite{Herth326}. The few feasibility studies \cite{fukuda2021safety} are clinical studies that collect this feasibility by doing a bronchoscopy and comparing the result with those of virtual bronchoscopy simulations.

The goal of this work is to analyze the feasibility of a guided bronchoscopic biopsy from a given POE through the  implementation of heatmaps in virtual bronchoscopy. %Each value of the heatmap represents the amount of extracted tissue given by the intersection between the uncertainty area of the guidance system and the lesion. 
%In order to make these computations, each structure (anatomical and guidance uncertainty area) has been represented as 3D objects. 
In particular, we contribute in the following aspects. Unlike existing VB planners, we compute a 3D scene of a guided biopsy that includes the patient's anatomy and the area of uncertainty caused by the error in the guidance system. This area is mathematically modelled as a cone that is defined from the elements of the anatomical scene and an error in the orientation of the guidance system. 
We model the feasibility of a biopsy taken from a given POE by means of intersections between the 3D objects of the biopsy scene. The feasibility  includes conditions to exclude POE not valid for a biopsy and a heatmap based on the amount of tissue that could be extracted from each valid POE given an error in guidance orientation. Finally, several configurations of the scene elements allow a statistical exploration of the impact on the viability of the biopsy of different morphological variables (like lesion shape and distance).

%The paper is organized as follows. In section \ref{amodel} we explain the calculation of the scene that models, both, the patient's anatomy and the uncertainty of the guidance using graphic primitives, in Section \ref{heatimplement} we describe the computation of feasibility of the biopsy using heatmaps. Experiments for the exploration of the impact of morphological variables are reported in \ref{sec:experiments} and, finally, Section \ref{sec:conclusions} concludes the paper. 

\section{Scene of the Biopsy}
\label{amodel}

Heatmaps are calculated from a scene that represents both the patient's lung anatomy and the uncertainty of the tunneling orientation. The scene of the anatomy is computed from two binary segmentations of the patient's airways and lesion. The scene of the orientation uncertainty is modelled using graphical primitives. Both scenes are modelled and visualized using the Python library Vedo \cite{musy2021vedo}. 

The input data to compute the anatomical scene are two binary volumes extracted from CTs representing the patient's nodule and bronchial anatomy. The volume of the nodule is a manual segmentation done by an expert using the 3DSlicer software. The volume of the bronchi has been obtained automatically with the segmentation algorithm described in \cite{gil2019segmentation}. These two volumes have been transformed into Vedo mesh objects. 
%by first creating a Vedo volume object from them using the Volume() Vedo function and then applying the Vedo volume method isosurface(). This generates a Vedo object mesh of the border voxels of the binary volume. 
%An important part of the preprocessing is the cleaning of the volumes, because both can contain multiple related components in addition to the bronchi and the nodule, for more advanced objectives all the components that do not correspond to these objects have been removed, associates from the noise
In order to represent all possible navigation paths, we extract the skeleton points of the bronchial volume. A 3D skeletonization is a process by which the central points or axes of a three-dimensional object, in this case the bronchi, are found. In our case, the skeleton is extracted from the airway's mesh using the Skeletoner library \cite{au2008skeleton}. 

%Figure\ref{fig1} shows a visualization of an example of anatomical scene, with the skeleton points of the semi-transparent model of airways in gray and an irregular lesion located in a distal bronchi. 

%\begin{figure}[t]
%\centerline{\includegraphics[width=15pc]{images/Anatomy.png}}
%\caption{Different anatomy meshes. Airway mesh in gray color; nodule %mesh in black color; and skeleton in doted lines.\label{fig1}}
%\end{figure}

\begin{figure}[!ht]
\centerline{\includegraphics[width=15pc]{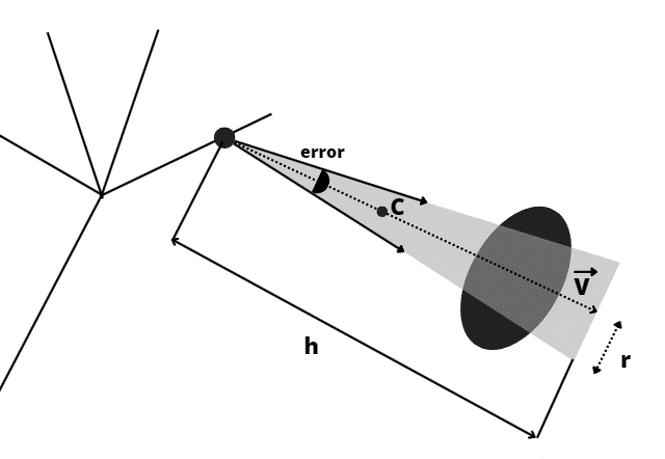}}
\caption{Graphical scheme for the interpretation of the uncertainty cone generation and its parameters.\label{fig2}}
\end{figure}

For the modelling of the uncertainty in tunneling orientation, we observe that, in an ideal guidance system, the orientation for biopsy tunneling instruments from a POE could be represented as a straight line in the direction of the lesion. A margin of error in this orientation generates an area of uncertainty around the central direction, which can be represented as a conical volume. Figure \ref{fig2} shows a  2D representation of a scene with the margin of error  given as the opening angle (labelled $error$ in the figure) of the triangle that crosses the lesion represented as a dark ellipse. In the 3D case, this angular margin can be oriented in any direction of the plane perpendicular to the height of the triangle   (given by the vector $\vec{V}$ in Figure \ref{fig2}) and, thus, this error generates a 3D cone. We define this 3D cone using the Cone class of Vedo, which generates the mesh of a cone from  parameters (see Figure \ref{fig2}) that determine its shape. The shape parameters are the height of the cone ($h$ in Figure \ref{fig2}), the direction vector ($\Vec{V}$ in Figure \ref{fig2}) defining the orientation of the cone, the center ($C$ in Figure \ref{fig2}) which is the middle point of the central axis of the cone and the radius of the base ($r$ in Figure \ref{fig2}) These parameters have been determined by studying the scene as we explain below. 

The height of the cone represents the distance that the instruments must travel in order to perform the biopsy. Since it must be large enough to completely cross the nodule, we defined it as the distance between the $POE$ and the farthest point of the nodule's bounding box. The bounding box is extracted with the GetBounds() method of the Vedo mesh objects.

The direction of the cone has been determined as the unit vector between $POE$ and the center of the nodule: 
\begin{equation}
\label{eq_0} \Vec{V} = \frac{POE-C_{Nod}}{|| POE-C_{Nod} ||_2}
\end{equation}

\noindent for $|| \cdot ||_2$ the Euclidean norm and the nodule center $C_{Nod}$ given by the GetCenter() method of the Vedo mesh object representing the lesion.

The center of the cone is the midpoint of the central axis. Given that this central axis is the line with origin $POE$ and vector director $\Vec{V}$,  the center is computed as:

\begin{equation}
\label{eq_1} C = POE-h /2 * \Vec{V}  
\end{equation}

Finally, the radius of the base of the cone, r, can be calculated from the angular uncertainty ($error$) of the guidance system using trigonometry (see \ref{fig2}) with the following formula:

\begin{equation}
\label{eq_2} 
r = h * \tan{(error)}
\end{equation}

\noindent where $tan(\cdot)$ is the tangent of an angle. 

%Figure\ref{fig3} shows all the elements involved in the computation of the biopsy scene. The airways (blue), the nodule (black), the 'bounding box' of the nodule (red), and the uncertainty cone (green).

%\begin{figure}[!ht]
%\centerline{\includegraphics[width=15pc]{images/elements.png}}
%\caption{Elements of the Guided Biopsy Scene. Anatomical: the airways (blue) and the nodule (black). Graphical: the 'bounding box' of the nodule (red), and the cone (green). \label{fig3}}
%\end{figure}

\section{Heatmap of the Biopsy Feasibility}
\label{heatimplement}

The computation of the heatmap has two main steps: selection of a set of valid POE and computation of the amount of tissue that could be extracted from them. 
Both steps are computed from intersections between the 3 objects of the biopsy scene, cone, lesion and airways. Intersections are computed in the voxel volumetric domain by adding the binary volumes of the masks associated to each 3D object. The value of this additive volume is equal to the number of intersecting objects: zero for background, and $N_{obj}>0$ for voxels belonging to $N_{obj}$ objects.

%\subsection{Candidate skeleton points}

\begin{figure}[!ht]
\centerline{\includegraphics[width=25pc]{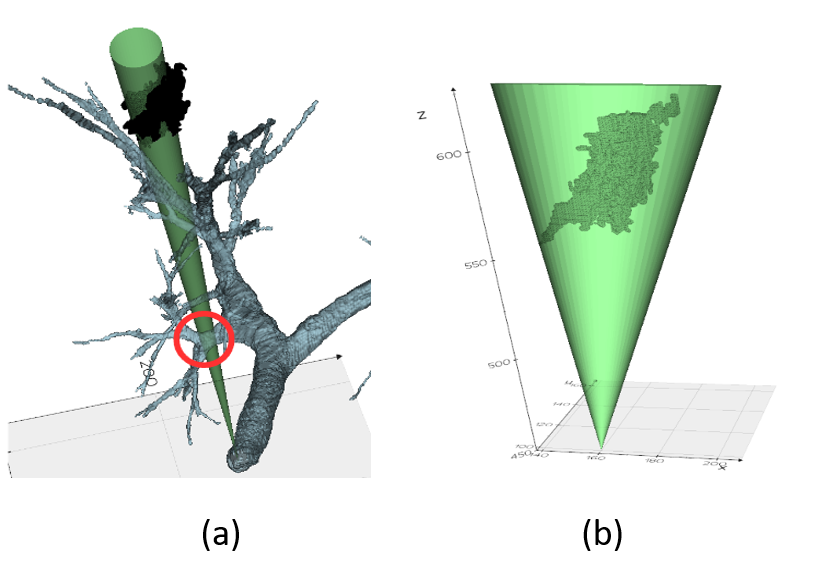}}
\caption{POE Feasibility Conditions: (a) Intersection between airway and cone uncertainty, marked with a red circle. (b) Nodule completely inside the cone uncertainty. \label{fig4}}
\end{figure}

There are two conditions on POE that make a biopsy from them infeasible. In case instruments cross other anatomic structures a POE should be discarded since it increases the risk for the patient. Also POE such that the cone of uncertainty includes surrounding lung tissue outside the nodule should be discarded since the sample could not contain enough lesion tissue. POE where any of these conditions occur are discarded. In order to determine whether instruments cross a bronchus, for each POE in the skeleton, we compute the intersection between the cone with vertex at this POE and the bronchial volume. If the maximum of the additive volume is greater than 1, then, the POE is discarded. Figure \ref{fig4} (a) shows an example of a POE invalid for the biopsy of the distal lesion with the bronchi traversed by its cone in a red circle. To discard POE's such with an uncertainty cone including surrounding lung tissue, we compute the intersection between the lesion volume and the cone boundary. The cone boundary is computed in the volumetric domain by subtracting the volumes of the original cone and a dilated version. POE's with the maximum of the additive intersection volume is less than 2 are discarded.

%The dilation is computed as the volumetric transformation of a Vedo cone with a radius 2 units larger than the radius of the uncertainty cone. If the maximum of the addition of the volume of the cone boundary and the lesion is less than two, it implies that the lesion is completely included inside the cone and the point is discarded. Figure\ref{fig4} (b) shows an example of a lesion completely included inside the uncertainty cone. 

%Those skeleton points such that none of the two conditions above are met are considered valid candidates for the biopsy of the lesion. We compute their feasibility from the intersection between the cone and the nodule volumes as follows. 

%\subsection{Heatmap Computation}

%\begin{figure*}[b]
%\centerline{\includegraphics[width=30pc]{images/distance.png}}
%\caption{Validity intersection scheme with respect to different distances from %airways to nodule. \label{fig5}}
%\end{figure*}

Heatmaps are computed from the maximum amount of tissue available for the biopsy from each valid POE, which corresponds to the number of voxels of the intersection volume with value equal to two. This quantity increases with the distance of the nodule to the POE. This is not consistent from a clinical point of view, since it has been shown that the more distant the lesion is, the more inconclusive the biopsy might be. We take this into account by the introduction of a distance penalty factor and formulate heatmaps at a valid $POE$ as the volume of the intersection POE cone and lesion divided by the distance from the center of the nodule, $C_{Nod}$ to $POE$:

\begin{equation}
heatmap(POE):=\frac{|V_{Nod} \bigcap V_{Cone}|}{||C_{Nod}-POE ||_2}
\end{equation}

\noindent where $V_{*}$ is the binary volume of a mesh object and $| \cdot|$ the volume of a 3D volumetric object.

\section{Experiments and Results}\label{sec:experiments}

The goal of these experiments is to explore the impact of the nodule shape, orientation and distance from bronchi on the feasibility of a biopsy. To do so, we have generated several lung scenes with two lesions with different shapes, one with a regular shperical shape and the other one with an irregular elongated and spiculated shape. Each nodule has been placed at 3 increasing distances (D1, D2, D3) from a distal point. In order to determine each distance, we extract a distance map from the airways mask to the rest of lung scene. Afterwards an histogram of the distances is applied to finally randomly select lung scene points (discarding such near outside the lungs) at different percentiles (25, 60, 90) representing distances. Finally we have located the nodule to each respective point, depending on the distance we were analyzing. 
For each distance, nodule has been rotated around its center. Each rotation is defined by the axis of rotation given by the unitary vector  $(cos(\theta)cos(\phi), sin(\theta)cos(\phi),sin(\phi))$, for $\theta \in [0º,180º]$, $\phi \in [0º,90º]$ and an angle of rotation $\alpha \in [0º,360º]$. The three ranges of the angles defining the rotation have been uniformly sampled every $45º$ to obtain 64 rotations, which gives a total number of 128 scenes (64 for each nodule shape). Regarding the anatomy of airways, we have used one of the binary volumes of the public CPAP database   
\cite{diez2018positive}. The CPAP dataBase contains CTs from 20 patients recorded in inspiration and expiration with 4 different positive airway pressure protocols. We have used the volume number (with size (321,464,767)) 14 in inspiration and standard pressure protocol for the definition of our anatomical scene. For each one of the 128 scenes, its heatmap has been computed in order to analyze the impact of the lesion shape and distance to the airways. The computational cost of the heatmap for each scene is around $42$ seconds on a computer of CPU: AMD Ryzen 7 5700U, GPU: AMD Radeon Vega 8 Graphics, and 32 GB of RAM.

In order to analyze the impact of the distance, Table \ref{tab1} reports a statistical summary of the distribution of heatmap values (median and IQR) for the 3 distances. For each distance, we report mean $\pm$ standard deviation of heatmap median and IQR computed for the 64 rotations. For the regular shape, the distribution of heatmap values have a significant drop in average values as distance increases, although the IQR remains stable. Whereas for the irregular shape, both, median and IQR, suffer a significant drop across increasing distances. The drop in ranges is attributed to a drop in the number of valid POE as the distance increases, which makes the distribution of values to be more concentrated around the median.

%The median is not greatly affected by the distance from the nodule. The IQR, which is equivalent to the size of the box, is indeed affected by the distance, the greater the distance the lower the mean of the IQR and the smaller the standard deviation of it, because the greater the distance the points that took high values reduce their value, due to the distance penalty, reducing the value that delimits the box, the points with low values are no longer candidates. This evolution means a significant drop in the median box size (IQR), and a reduction in the standard deviation, small boxes stay small and large boxes decrease in size (less variability).

\begin{table}[h]%
\centering
\caption{Impact of rotation across distance. Ranges of medians and IQR on heatmap distribution.\label{tab1}}%
\begin{tabular*}{240pt}{@{\extracolsep\fill}lccc@{\extracolsep\fill}}%
\toprule
 & \textbf{D1} & \textbf{D2} & \textbf{D3} \\
\midrule
\textbf{Reg. Med.} & $27.94\pm0.69$ & $22.78\pm0.92$ & $19.62\pm0.84$ \\
\textbf{Irr. Med.} & $15.47\pm1.30$ & $13.05\pm1.87$ & $11.82\pm1.79$ \\
\textbf{Reg. IQR} & $6.40\pm0.99$ & $7.69\pm1.07$ & $6.78\pm0.73$ \\
\textbf{Irr. IQR} & $4.85\pm1.60$ & $3.96\pm1.38$ & $3.31\pm1.62$ \\
\bottomrule
\end{tabular*}
\end{table}

\begin{figure}[h]
\centerline{\includegraphics[width=30pc]{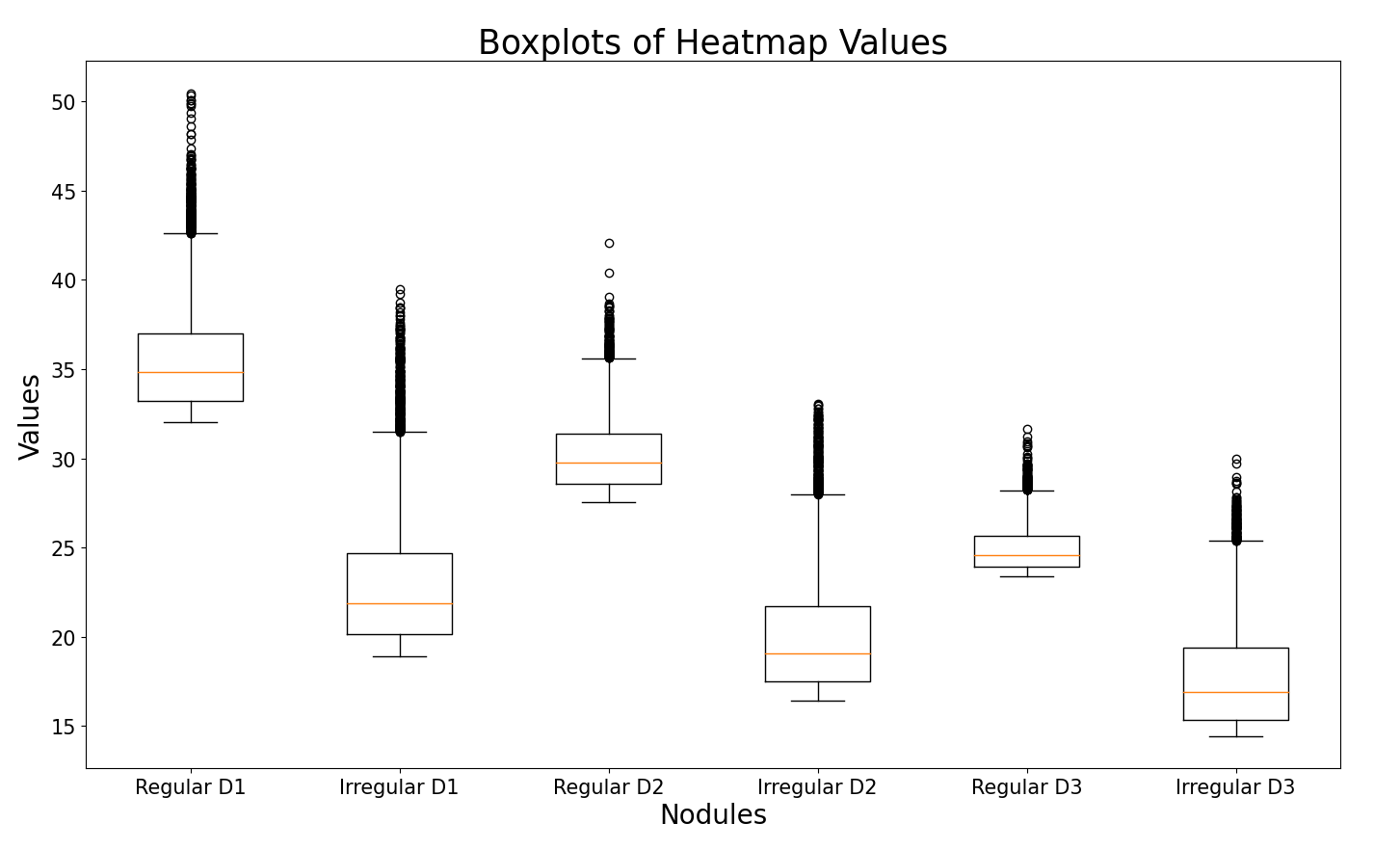}}
\caption{Boxplots assessing the impact of shape across distance. \label{fig8}}
\end{figure}

Comparing  ranges across distances, as expected, ranges decrease as distance increases. This decrease is more significant in the case of a regular shape, specially for the POE with highest quality. The boxplots in Figure \ref{fig8} illustrate these differences between values by shape and distance for the $20\%$ of the highest values. 

\begin{figure}[!ht]
\centerline{\includegraphics[width=35pc]{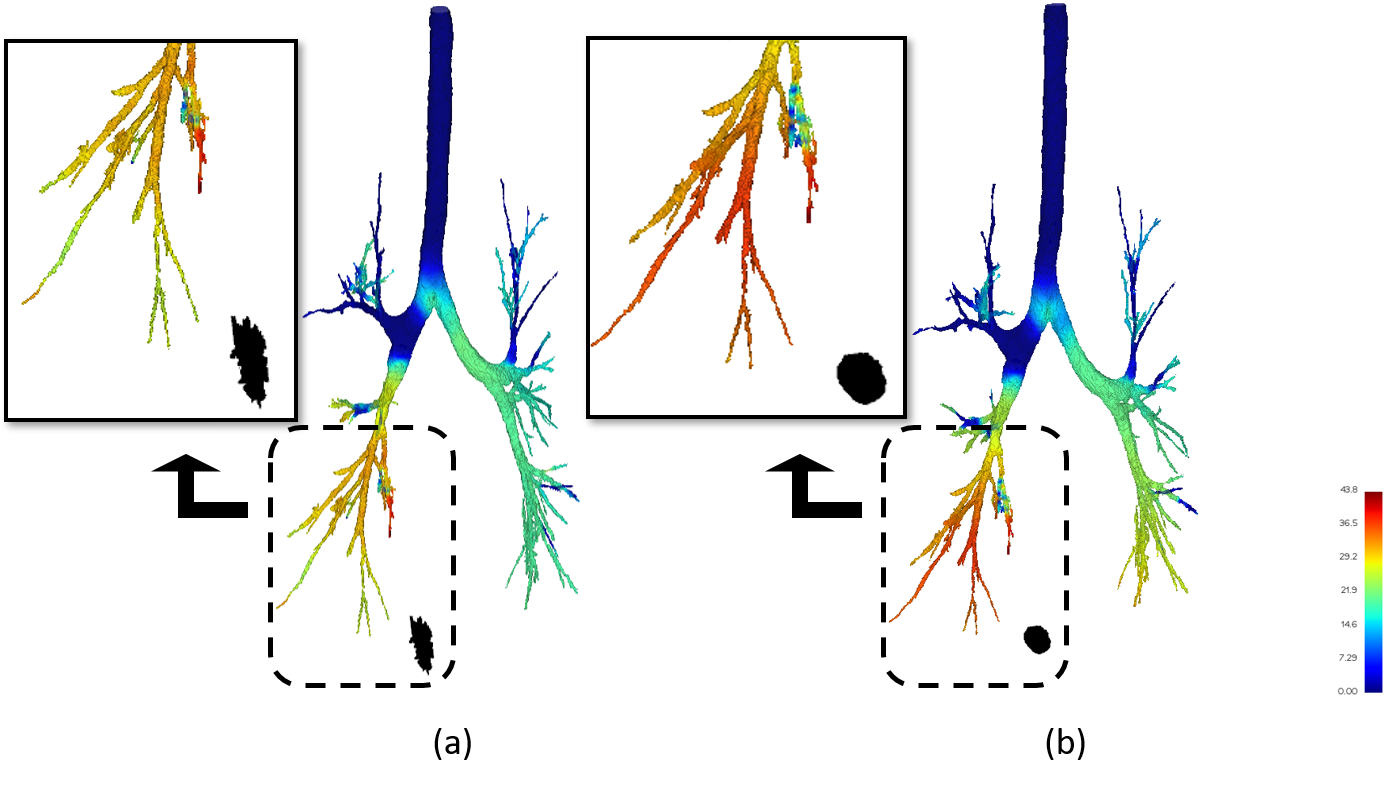}}
\caption{Visualization of heatmaps for an irregular, (a), and regular, (b) nodules. \label{fig6}}
\end{figure}

Figure \ref{fig6} shows the visualizations of two heatmaps, the one on the left computed for an irregular (malign) lesion and the one on the right computed for a regular (benign) nodule. For the one with the irregular lesion it can be seen that the hottest point it is not the closest point, but the one with the cone that fits you best in terms of alignment with the orientation of the lesion. It demonstrates that orientation of irregular nodules is a key point in POE selection.

\section{Conclusions}\label{sec:conclusions}

%This paper has presented a virtual tool for the analysis of the feasibility of a guided biopsy taking into account potential errors in the orientation of the guiding system. We have used Python Vedo graphical library to model a virtual scene of the guided biopsy including anatomy and uncertainty in the orientation of the biopsy instruments. The intersection between the elements of this biopsy scene allow the computation of the amount of tissue that could be biopsed from a set of valid POE. This defines our feasibility heatmap. Several configurations of these virtual scenes allow the statistical analysis of the impact of structural variables, lesion shape and distance to bronchi in our experiments. 

As far as we know this is the first work that addresses the computation of the feasibility of a guided biopsy using virtual bronchoscopy.  We consider that virtual bronchoscopy systems could benefit in several aspects if they incorporated heatmap analysis for POE location. These heatmaps could help to determine which is the optimal POE, model the maximum error allowed in the navigation systems, as well as study which variables have a greater influence on the success of the operation. The statistical analysis of the impact of the variables has allowed a better understanding of how the different morphological factors influence the biopsy yield and can provide useful information for decision-making in future interventions. 
Virtual bronchoscopy softwares are widely accessible in most centers, and the incorporation of heatmaps could represent a low-cost upgrade with significant potential impact on diagnosing extrabronchial lesions. This could be particularly valuable for less complex.
Furthermore, being able to study the feasibility of the intervention in terms of morphological and structural anatomical variables could help to discard interventions having a high probability of extracting a sample of a sub-optimal quality for a conclusive biopsy. Future steps will include comparison with existing pacification systems as well as incorporating other variables having an impact in the quality of a biopsy like instruments physics or nodule density.

\section*{Declarations}

\begin{itemize}

\item \textbf{Funding:} This study was funded by the Ministerio de Economía, Industria y Competitividad, Gobierno de España grant number  PID 2021-126776OB-C21, Agència de Gestió d'Ajuts Universitaris i de Recerca grant numbers 2021SGR01623 and CERCA Programme / Generalitat de Catalunya.

\item \textbf{Conflict of interest:} The authors declare that they have no conflict of interest.

\item \textbf{Ethics approval:} For this type of study formal consent is not required.

\end{itemize}

\bibliography{biblioCARS2024}% common bib file

%% BioMed_Central_Bib_Style_v1.01

\begin{thebibliography}{12}
% BibTex style file: bmc-mathphys.bst (version 2.1), 2014-07-24
\ifx \bisbn   \undefined \def \bisbn  #1{ISBN #1}\fi
\ifx \binits  \undefined \def \binits#1{#1}\fi
\ifx \bauthor  \undefined \def \bauthor#1{#1}\fi
\ifx \batitle  \undefined \def \batitle#1{#1}\fi
\ifx \bjtitle  \undefined \def \bjtitle#1{#1}\fi
\ifx \bvolume  \undefined \def \bvolume#1{\textbf{#1}}\fi
\ifx \byear  \undefined \def \byear#1{#1}\fi
\ifx \bissue  \undefined \def \bissue#1{#1}\fi
\ifx \bfpage  \undefined \def \bfpage#1{#1}\fi
\ifx \blpage  \undefined \def \blpage #1{#1}\fi
\ifx \burl  \undefined \def \burl#1{\textsf{#1}}\fi
\ifx \doiurl  \undefined \def \doiurl#1{\url{https://doi.org/#1}}\fi
\ifx \betal  \undefined \def \betal{\textit{et al.}}\fi
\ifx \binstitute  \undefined \def \binstitute#1{#1}\fi
\ifx \binstitutionaled  \undefined \def \binstitutionaled#1{#1}\fi
\ifx \bctitle  \undefined \def \bctitle#1{#1}\fi
\ifx \beditor  \undefined \def \beditor#1{#1}\fi
\ifx \bpublisher  \undefined \def \bpublisher#1{#1}\fi
\ifx \bbtitle  \undefined \def \bbtitle#1{#1}\fi
\ifx \bedition  \undefined \def \bedition#1{#1}\fi
\ifx \bseriesno  \undefined \def \bseriesno#1{#1}\fi
\ifx \blocation  \undefined \def \blocation#1{#1}\fi
\ifx \bsertitle  \undefined \def \bsertitle#1{#1}\fi
\ifx \bsnm \undefined \def \bsnm#1{#1}\fi
\ifx \bsuffix \undefined \def \bsuffix#1{#1}\fi
\ifx \bparticle \undefined \def \bparticle#1{#1}\fi
\ifx \barticle \undefined \def \barticle#1{#1}\fi
\bibcommenthead
\ifx \bconfdate \undefined \def \bconfdate #1{#1}\fi
\ifx \botherref \undefined \def \botherref #1{#1}\fi
\ifx \url \undefined \def \url#1{\textsf{#1}}\fi
\ifx \bchapter \undefined \def \bchapter#1{#1}\fi
\ifx \bbook \undefined \def \bbook#1{#1}\fi
\ifx \bcomment \undefined \def \bcomment#1{#1}\fi
\ifx \oauthor \undefined \def \oauthor#1{#1}\fi
\ifx \citeauthoryear \undefined \def \citeauthoryear#1{#1}\fi
\ifx \endbibitem  \undefined \def \endbibitem {}\fi
\ifx \bconflocation  \undefined \def \bconflocation#1{#1}\fi
\ifx \arxivurl  \undefined \def \arxivurl#1{\textsf{#1}}\fi
\csname PreBibitemsHook\endcsname

%%% 1
\bibitem[\protect\citeauthoryear{Nadig et~al.}{2023}]{nadig2023guided}
\begin{barticle}
\bauthor{\bsnm{Nadig}, \binits{T.R.}},
\bauthor{\bsnm{Thomas}, \binits{N.}},
\bauthor{\bsnm{Nietert}, \binits{P.J.}},
\bauthor{\bsnm{Lozier}, \binits{J.}},
\bauthor{\bsnm{Tanner}, \binits{N.T.}},
\bauthor{\bsnm{Memoli}, \binits{J.S.W.}},
\bauthor{\bsnm{Pastis}, \binits{N.J.}},
\bauthor{\bsnm{Silvestri}, \binits{G.A.}}:
\batitle{Guided bronchoscopy for the evaluation of pulmonary lesions: An updated meta-analysis}.
\bjtitle{Chest}
\bvolume{163}(\bissue{6}),
\bfpage{1589}--\blpage{1598}
(\byear{2023})
\end{barticle}
\endbibitem

%%% 2
\bibitem[\protect\citeauthoryear{Shepherd et~al.}{}]{shepherd2018image}
\begin{botherref}
\oauthor{\bsnm{Shepherd}, \binits{W.}},
\oauthor{\bsnm{Colt}, \binits{H.}},
\oauthor{\bsnm{Finlay}, \binits{G.}}:
Image-guided bronchoscopy for biopsy of peripheral pulmonary lesions.
UpToDate (Sept. 2018).
\url{https://www.uptodate.com/contents/image-guided-bronchoscopy-for-biopsy-of-peripheral-pulmonary-lesions/print}
\end{botherref}
\endbibitem

%%% 3
\bibitem[\protect\citeauthoryear{Ramzy et~al.}{2018}]{ramzy2018biopsy}
\begin{botherref}
\oauthor{\bsnm{Ramzy}, \binits{J.}},
\oauthor{\bsnm{Travaline}, \binits{J.}},
\oauthor{\bsnm{Thomas}, \binits{J.}},
\oauthor{\bsnm{Basile}, \binits{M.}},
\oauthor{\bsnm{Massetti}, \binits{P.}},
\oauthor{\bsnm{Criner}, \binits{G.}}:
Biopsy through lung parenchymal lesion using virtual bronchoscopy navigation (VBN) Archimedes with EBUS sheath tunneling.
Eur Respiratory Soc
(2018)
\end{botherref}
\endbibitem

%%% 4
\bibitem[\protect\citeauthoryear{Burks and Akulian}{2022}]{burks2022electromagnetic}
\begin{bchapter}
\bauthor{\bsnm{Burks}, \binits{A.C.}},
\bauthor{\bsnm{Akulian}, \binits{J.}}:
\bctitle{Electromagnetic Navigation Bronchoscopy}.
In: \beditor{\bsnm{Wahidi}, \binits{M.M.}},
\beditor{\bsnm{Ost}, \binits{D.E.}} (eds.)
\bbtitle{Practical Guide to Interventional Pulmonology},
pp. \bfpage{23}--\blpage{33}.
\bpublisher{Elsevier Health Sciences},
\blocation{E-book}
(\byear{2022})
\end{bchapter}
\endbibitem

%%% 5
\bibitem[\protect\citeauthoryear{Avasarala et~al.}{2022}]{avasarala2022sight}
\begin{barticle}
\bauthor{\bsnm{Avasarala}, \binits{S.K.}},
\bauthor{\bsnm{Roller}, \binits{L.}},
\bauthor{\bsnm{Katsis}, \binits{J.}},
\bauthor{\bsnm{Chen}, \binits{H.}},
\bauthor{\bsnm{Lentz}, \binits{R.J.}},
\bauthor{\bsnm{Rickman}, \binits{O.B.}},
\bauthor{\bsnm{Maldonado}, \binits{F.}}:
\batitle{Sight unseen: diagnostic yield and safety outcomes of a novel multimodality navigation bronchoscopy platform with real-time target acquisition}.
\bjtitle{Respiration}
\bvolume{101}(\bissue{2}),
\bfpage{166}--\blpage{173}
(\byear{2022})
\end{barticle}
\endbibitem

%%% 6
\bibitem[\protect\citeauthoryear{Diez-Ferrer et~al.}{2019}]{diez2019ultrathin}
\begin{barticle}
\bauthor{\bsnm{Diez-Ferrer}, \binits{M.}},
\bauthor{\bsnm{Morales}, \binits{A.}},
\bauthor{\bsnm{Teb{\'e}}, \binits{C.}},
\bauthor{\bsnm{Cubero}, \binits{N.}},
\bauthor{\bsnm{L{\'o}pez-Lisbona}, \binits{R.}},
\bauthor{\bsnm{Padrones}, \binits{S.}},
\bauthor{\bsnm{Aso}, \binits{S.}},
\bauthor{\bsnm{Dorca}, \binits{J.}},
\bauthor{\bsnm{Gil}, \binits{D.}},
\bauthor{\bsnm{Rosell}, \binits{A.}}:
\batitle{Ultrathin bronchoscopy with and without virtual bronchoscopic navigation: influence of segmentation on diagnostic yield}.
\bjtitle{Respiration}
\bvolume{97}(\bissue{3}),
\bfpage{252}--\blpage{258}
(\byear{2019})
\end{barticle}
\endbibitem

%%% 7
\bibitem[\protect\citeauthoryear{Herth et~al.}{2015}]{Herth326}
\begin{barticle}
\bauthor{\bsnm{Herth}, \binits{F.J.}},
\bauthor{\bsnm{Eberhardt}, \binits{R.}},
\bauthor{\bsnm{all.}}:
\batitle{Bronchoscopic transparenchymal nodule access (btpna): first in human trial of a novel procedure for sampling solitary pulmonary nodules}.
\bjtitle{Thorax}
\bvolume{70}(\bissue{4}),
\bfpage{326}--\blpage{332}
(\byear{2015})
\end{barticle}
\endbibitem

%%% 8
\bibitem[\protect\citeauthoryear{Fukuda et~al.}{2021}]{fukuda2021safety}
\begin{barticle}
\bauthor{\bsnm{Fukuda}, \binits{Y.}},
\bauthor{\bsnm{Sugimoto}, \binits{H.}},
\bauthor{\bsnm{Yamada}, \binits{Y.}},
\bauthor{\bsnm{Ito}, \binits{H.}},
\bauthor{\bsnm{Tanaka}, \binits{T.}},
\bauthor{\bsnm{Yoshida}, \binits{T.}},
\bauthor{\bsnm{Okamori}, \binits{S.}},
\bauthor{\bsnm{Ando}, \binits{K.}},
\bauthor{\bsnm{Okada}, \binits{Y.}}:
\batitle{Safety and feasibility of lung biopsy in diagnosis of acute respiratory distress syndrome: protocol for a systematic review and meta-analysis}.
\bjtitle{BMJ open}
\bvolume{11}(\bissue{2}),
\bfpage{043600}
(\byear{2021})
\end{barticle}
\endbibitem

%%% 9
\bibitem[\protect\citeauthoryear{Musy et~al.}{2021}]{musy2021vedo}
\begin{botherref}
\oauthor{\bsnm{Musy}, \binits{M.}},
\oauthor{\bsnm{Jacquenot}, \binits{G.}},
\oauthor{\bsnm{Dalmasso}, \binits{G.}},
\oauthor{\bsnm{Bruin}, \binits{R.}},
\oauthor{\bsnm{Pollack}, \binits{A.}},
\oauthor{\bsnm{Claudi}, \binits{F.}},
\oauthor{\bsnm{Badger}, \binits{C.}},
\oauthor{\bsnm{Sullivan}, \binits{B.}},
\oauthor{\bsnm{Hrisca}, \binits{D.}},
\oauthor{\bsnm{Volpatto}, \binits{D.}}:
vedo: A python module for scientific analysis and visualization of 3d objects and point clouds.
Zenodo
(2021)
\end{botherref}
\endbibitem

%%% 10
\bibitem[\protect\citeauthoryear{Gil et~al.}{2019}]{gil2019segmentation}
\begin{barticle}
\bauthor{\bsnm{Gil}, \binits{D.}},
\bauthor{\bsnm{Sanchez}, \binits{C.}},
\bauthor{\bsnm{Borras}, \binits{A.}},
\bauthor{\bsnm{Diez-Ferrer}, \binits{M.}},
\bauthor{\bsnm{Rosell}, \binits{A.}}:
\batitle{Segmentation of distal airways using structural analysis}.
\bjtitle{Plos one}
\bvolume{14}(\bissue{12}),
\bfpage{0226006}
(\byear{2019})
\end{barticle}
\endbibitem

%%% 11
\bibitem[\protect\citeauthoryear{Au et~al.}{2008}]{au2008skeleton}
\begin{barticle}
\bauthor{\bsnm{Au}, \binits{O.K.-C.}},
\bauthor{\bsnm{Tai}, \binits{C.-L.}},
\bauthor{\bsnm{Chu}, \binits{H.-K.}},
\bauthor{\bsnm{Cohen-Or}, \binits{D.}},
\bauthor{\bsnm{Lee}, \binits{T.-Y.}}:
\batitle{Skeleton extraction by mesh contraction}.
\bjtitle{ACM transactions on graphics (TOG)}
\bvolume{27}(\bissue{3}),
\bfpage{1}--\blpage{10}
(\byear{2008})
\end{barticle}
\endbibitem

%%% 12
\bibitem[\protect\citeauthoryear{Diez-Ferrer et~al.}{2018}]{diez2018positive}
\begin{barticle}
\bauthor{\bsnm{Diez-Ferrer}, \binits{M.}},
\bauthor{\bsnm{Gil}, \binits{D.}},
\bauthor{\bsnm{Tebe}, \binits{C.}},
\bauthor{\bsnm{Sanchez}, \binits{C.}},
\bauthor{\bsnm{Cubero}, \binits{N.}},
\bauthor{\bsnm{L{\'o}pez-Lisbona}, \binits{R.}},
\bauthor{\bsnm{Dorca}, \binits{J.}},
\bauthor{\bsnm{Rosell}, \binits{A.}},
\bauthor{\bsnm{group}, \binits{P.-e.C.}}:
\batitle{Positive airway pressure to enhance computed tomography imaging for airway segmentation for virtual bronchoscopic navigation}.
\bjtitle{Respiration}
\bvolume{96}(\bissue{6}),
\bfpage{525}--\blpage{534}
(\byear{2018})
\end{barticle}
\endbibitem

\end{thebibliography}
%% if required, the content of .bbl file can be included here once bbl is generated
%\input sn-article.bbl

\end{document}